*Review Article*

# Modern Machine-Learning Predictive Models for Diagnosing Infectious Diseases

**Eman Yahia Alqaissi** ,[1,2] **Fahd Saleh Alotaibi** ,[1] **and Muhammad Sher Ramzan** [1]

[1]*Information Systems, Faculty of Computing and Information Technology, King Abdulaziz University, Jeddah, Saudi Arabia*
[2]*Information Systems, King Khalid University, Abha, Saudi Arabia*

Correspondence should be addressed to Eman Yahia Alqaissi; eabdoalqaissi@stu.kau.edu.sa

Controlling infectious diseases is a major health priority because they can spread and infect humans, thus evolving into epidemics or pandemics. Therefore, early detection of infectious diseases is a significant need, and many researchers have developed models to diagnose them in the early stages. This paper reviewed research articles for recent machine-learning (ML) algorithms applied to infectious disease diagnosis. We searched the Web of Science, ScienceDirect, PubMed, Springer, and IEEE databases from 2015 to 2022, identified the pros and cons of the reviewed ML models, and discussed the possible recommendations to advance the studies in this field. We found that most of the articles used small datasets, and few of them used real-time data. Our results demonstrated that a suitable ML technique depends on the nature of the dataset and the desired goal. Moreover, heterogeneous data could ensure the model's generalization, while big data, many features, and a hybrid model will increase the resulting performance. Furthermore, using other techniques such as deep learning and NLP to extract vast features from unstructured data is a powerful approach to enhancing the performance of ML diagnostic models.

## 1. Introduction

Infectious diseases constitute major health issues. Viruses, fungi, bacteria, and parasites cause infectious diseases, which are transmitted from infected humans or a pathogen of an animal to other humans. The World Health Organization (WHO) has reported that infectious diseases, including lower respiratory infections, malaria, tuberculosis, and HIV/AIDS, were among the top ten leading causes of death worldwide in 2019 [1].

Several applied techniques focused on developing models for classifying different infectious diseases. In recent years, infectious disease outbreaks were still a common problem worldwide. Therefore, early detection and diagnosis of infectious diseases are critical to preventing and controlling them efficiently. According to Goodman et al. [2], timely and accurate diagnosis of infectious diseases is vital to managing the disease correctly.

Artificial intelligence (AI) is used to classify and predict the spread of infectious diseases [3]. AI assists computers in performing tasks that require human intelligence. John McCarthy introduced the term "AI" in 1956. Still, studies on machine ways of thinking were published before this time, with Vannevar Bush's seminal work and Alan Turing's paper in 1950 about machines' ability to think and simulate human work [4].

Machine-learning (ML) is a branch of AI that learns from data and makes a prediction. ML algorithms come in three main types: supervised, unsupervised, and reinforcement learning. In supervised learning, a model is trained on given data and independent variables to predict the dependent variable and then achieve the desired accuracy, such as the decision tree, random forest (RF), logistic regression (LR), and $k$-nearest neighbors. On the other hand, unsupervised learning identifies patterns in training data that are not classified or labeled, then categorizes them based on the extracted features, such as the Apriori algorithm and $k$-means. Moreover, the reinforcement learning model trains a machine to learn from experience and make an accurate decision through trial and error.

This review paper is interested in using ML to assist in the diagnosis of several infectious diseases and answers the following research questions:

(i) What ML techniques are used for diagnosing infectious diseases?

(ii) What are the types of datasets used in the diagnostic models?

(iii) What are the performance metrics used for infectious disease diagnosis models?

The large number of techniques that are emerging makes it necessary to provide a precise overview of diagnosing infectious diseases. To the best of our knowledge, this is the first review to investigate existing works on diagnosing, detecting, and classifying infectious diseases to answer these questions and thereby assist with new and precise ML techniques for detecting infectious diseases early. Furthermore, early and accurate detection of infectious diseases plays an important role in future possible outbreak prevention. In addition, we provide trends in diagnosing infectious diseases using ML algorithms and future research directions.

Most of the previous works demonstrated that ML approaches were adopted by many researchers, but there is no review to examine them deeply. To that end, we propose a systematic review dealing with the ML models applied to the diagnosis of infectious diseases. We discuss and describe the dataset, utilized technique, and performance for each model. Based on the obtained explanations and discussion, we provide recommendations for future research and trends in the area to assist in creating an accurate and perhaps generalized model for detecting infectious diseases early to avoid possible outbreaks.

We organized the structure of this review as follows. First, the related work section covers reviews and systematic review papers for infectious disease classification and diagnosis models. Then, in the search strategy section, we define the criteria for the included articles. After that, we provide a detailed explanation of every article in the result section. Then, we present a discussion and future research trends. The final section is the conclusion.

## 2. Related Work

In the literature, several systematic review papers introduced the use of ML in infectious disease classification. However, these papers are concerned with reviewing ML diagnoses for only specific infectious diseases such as COVID-19 [5, 6], hepatitis [7], pneumonia [8], tuberculosis [9–11], and influenza A virus phenotypes [12].

Other papers are concerned with several infectious diseases, but these reviews specified applications of digital technologies, artificial intelligence, and machine-learning in infectious disease laboratory testing to overcome human analytical limitations [13, 14]. Other reviews focused on all kinds of infections and noninfectious diseases [15, 16].

Table 1 summarizes the main characteristics of related works to differentiate between them and our systematic review paper.

## 3. Search Strategy

We rely on the process of systematic review preferred reporting items for systematic review and meta-analysis (PRISMA) [17] to collect, review, and answer the research questions. It has three main stages: identification, screening process, and included studies, as shown in Figure 1.

3.1. Identification and Search Sources. This paper explored four databases: Web of Science, PubMed, Springer, IEEE, and ScienceDirect, to find related research articles from 2015 to 2022. The combination of keywords used for search queries is as follows: "classification model for infectious disease," "ML for diagnosing infectious diseases," and "artificial intelligence for infectious diseases' classification." The search yielded 9490 English language records, 4241 from the Web of Science, 1771 from PubMed, 3299 from ScienceDirect, 147 from IEEE, and 32 from Springer. In addition, we used an automation tool to remove 2168 duplicated records and the 56 non-English language articles.

3.2. Screening Process. This study focuses only on ML algorithms for human studies. It categorizes retrieved diagnosis models into six categories based on the applied dataset. We have models that used signs and symptoms, models that used image processing, models that used clinical tests, models that used clinical reports, models that used electronic health records, and models that used a combination of several predictors.

This study excludes 2364 ineligible articles. Of these, 2148 were irrelevant, and 216 were literature and systematic reviews. The remaining articles were 4902; eliminated articles were 185 as they provide diagnosis models about infectious diseases besides other noninfectious diseases and 1456 articles about transmission and spread models of infectious disease. Both are not related to this review. The final retrieved articles were 3261.

3.3. Included Studies. This review screened each of the 3261 retrieved full-text articles by the authors of this study through the systematic review accelerator tool. We used the checklist for critical appraisal and data extraction for systematic reviews of prediction modeling studies (CHARMS) [18] to assess these retrieved articles. The abstract was reviewed with the CHARMS checklist, as illustrated in Table 2. We found 1091 articles based on other techniques such as graph models, infrared spectroscopy, risk detection models, and severity assessment models. In addition, 115 articles focused on deep learning techniques, which we did not cover in this review. In addition, we excluded animal studies (356 articles), models for predictive therapy and drugs (819 articles), and studies about infectious diseases' genes, genome, sequencing, and other analysis methods (866 articles). Finally, we found only 14 articles that matched our inclusion criteria.

TABLE 1: Comparison of related systematic reviews and our systematic review.

| Reference | Focus | Differences |
|---|---|---|
| [5] | (i) Diagnosing COVID-19 and predicting severity and mortality risks | (i) The review is based only on clinical and laboratory data. |
| [6] | (i) Diagnosis and prognosis of COVID-19 from prediction models. | (i) The review focuses more on preprints<br>(ii) It covers all types of models including risk prediction, diagnosis of severity, and diagnosis from images |
| [7] | (i) Diagnosing hepatitis | (i) It covers only clinical tests. |
| [8] | (i) Detecting pneumonia | (i) It is based only on signs and symptoms<br>(ii) Performance measures are not covered |
| [9] | (i) Diagnosing tuberculosis | (i) It covers diverse AI approaches using clinical signs and symptoms and radiological images. |
| [10] | (i) Diagnosing pulmonary tuberculosis | (i) It covers AI methods based on chest X-ray images. |
| [11] | (i) Diagnosing tuberculous meningitis | (i) It is based only on clinical and laboratory data. |
| [12] | (i) Predicting phenotypic characteristics of influenza virus | (i) It is based on genomic or proteomic input. |
| [13] | (i) Diagnosing HIV, HCV, and chlamydia | (i) It implements different digital technology but does not include any kind of AI technique. |
| [14] | (i) Diagnosing COVID-19, hepatitis, sepsis, malaria, Lyme disease, and tuberculosis | (i) It covers data coming from EMR. |
| [15] | (i) Automatic diagnosis of several infections such as sepsis, general infections, and *Clostridium difficile* infection through ML and expert system | (i) It covers papers based on physiological data. |
| [16] | (i) Diagnosing infectious and noninfectious diseases through ML | (i) It explains in detail all reviewed ML algorithms but does not mention datasets or performance measures. |
| Our review | (i) ML diagnosis of all available human infectious disease papers | (i) It covers different kinds of ML techniques, several types of datasets, and performance measures. |

HIV: human immunodeficiency virus; HCV: hepatitis C virus; EMR: electronic medical record.

## 4. Search Results

This paper reviewed each selected article descriptively. In the upcoming subsections, we covered explanations of datasets, types of ML techniques for each study, and the models' performance, such as precision, accuracy, specificity, and sensitivity.

*4.1. Dataset Specification.* The most retrieved ML diagnosis models are based on signs and symptoms [19–24], followed by models based on laboratory tests [25, 26]. Moreover, we found ML models based on clinical and electronic health records (EHRs) [27, 28], ML models based on clinical reports [29], ML models based on image processing with other techniques [30, 31], and ML models based on a combination of predictors including abnormal lab test results, the incidence rates, and signs and symptoms, as well as epidemiological features [32]. Figure 2 summarizes these different acquisition methods. Moreover, Table 3 shows the sizes, acquisition methods, and type of infectious disease of each article.

*4.2. Dataset Features.* Tetanus and HFMD cause autonomic nervous system dysfunction (ANSD) and lead to death in complicated stages. Moreover, physiological waveforms collected from sensors are susceptible to noise and should be filtered using a pass filter followed by a Gaussian filter [19]. In contrast, the biosensors in [20] collected five vital signs directly from patients to a hub that quantifies and mul-tiplexes the signals and sends them to a mobile application through Bluetooth technology.

On the other hand, a smartphone application was adopted [21] to collect data available in questionnaire format from registered users. However, the study [22] compared real-time polymerase chain reaction (RT-PCR) results with the diagnosis results of vocal inputs at the time of recordings and the required dataset available in WAV formats.

Twitter application program interfaces (APIs) are used in [23] to extract data from users' messages and then implement preprocessing to enhance quality and remove noise. Moreover, real electronic health records for 104 individuals diagnosed with influenza are used to validate the obtained results.

Domain ontology is constructed in [24] to diagnose 507 different infectious diseases based on the signs and symptoms of patients. This ontology-based model is more complete than the previous work in the field and performs data-driven decision-making at the point of primary care. On the other hand, the emergency department (ED) free-text reports were implemented to detect influenza cases based on 31 clinical findings adopted for ML classifiers [25].

Moreover, combining routine laboratory tests with ML models [26, 27] provides precious opportunities to assist in infectious disease diagnosis. The selected tests in [27] include six clinical chemistry (CC) parameters and 14 complete blood count (CBC) parameters. Moreover, 35 features were used to diagnose COVID-19 in [26], and random sampling was done for 22,385 patients with at least 33 out of 35 total features.

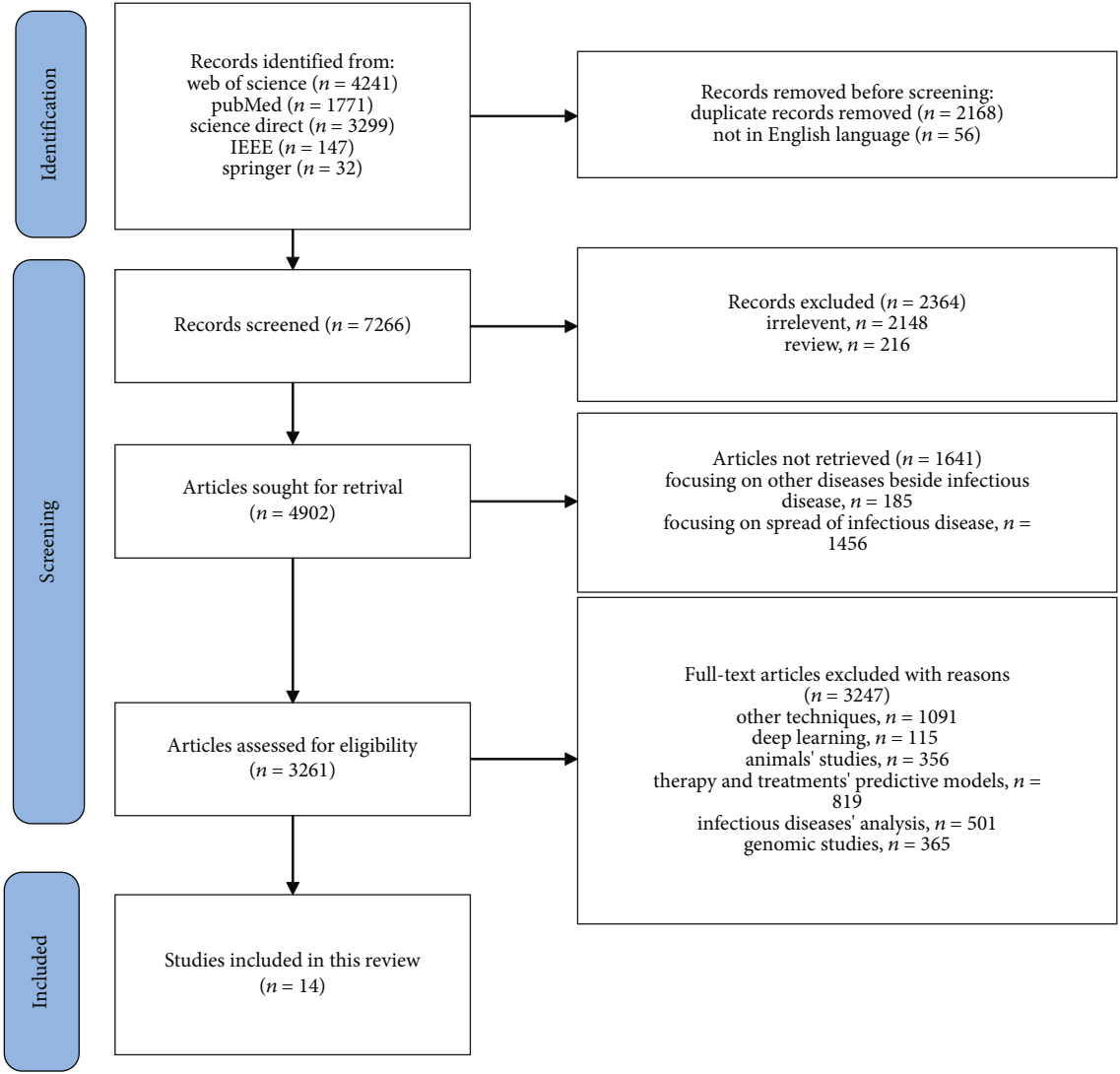

FIGURE 1: PRISMA flow diagram.

TABLE 2: CHARMS checklist for abstracts in the selected articles.

| Criteria | Description |
| --- | --- |
| "Type of diagnosis model" | Only ML algorithms classify infectious diseases |
| "Type of dataset" | Dataset specified and declared |
| "Evaluation of the model" | The performance metrics of the ML model |

Furthermore, a patient record system is a clinical information system that provides valuable information, including disease diagnosis. Hence, it assists with patient care. Clinical records [28] and EHR [29] were used to collect demographic and clinical data of patients, aiding in infectious disease diagnosis.

Feature engineering is a critical task for classification, and it is helpful to extract features from computerized tomography (CT) scan images. A novel convolutional neural network (CNN) model was applied [31] to assist with ML classification. Moreover, the authors of the study [31] applied contrast limited histogram equalization (CLAHE) to enhance the quality of the images. However, various datasets, including clinical data (demographics, radiological scores, and laboratory tests) and lesion/lung radiomic features extracted from enhanced chest CT images, were implemented in [30], to diagnose COVID-19.

In addition, a classifier was established in [32] to classify 25 common infectious diseases. It used symptoms and signs, city of disease origin, epidemiological features of the disease, and abnormal lab test results as input, but the authors did not specify all features that the model used. Table 4 lists each dataset's features.

*4.3. ML Algorithms and Model Performance.* Different ML algorithms in the reviewed articles focused on the nature of the data and the goal to be achieved [33]. Further, different performance metrics are applied to evaluate different ML algorithms. The right choice of metrics [34] influences the model's effectiveness based on test datasets and how to measure and compare the performance of ML algorithms. These metrics include accuracy, recall (sensitivity), precision, specificity, and F1-score. Figure 3 classifies reviewed articles according to the ML algorithms that each model adopted.

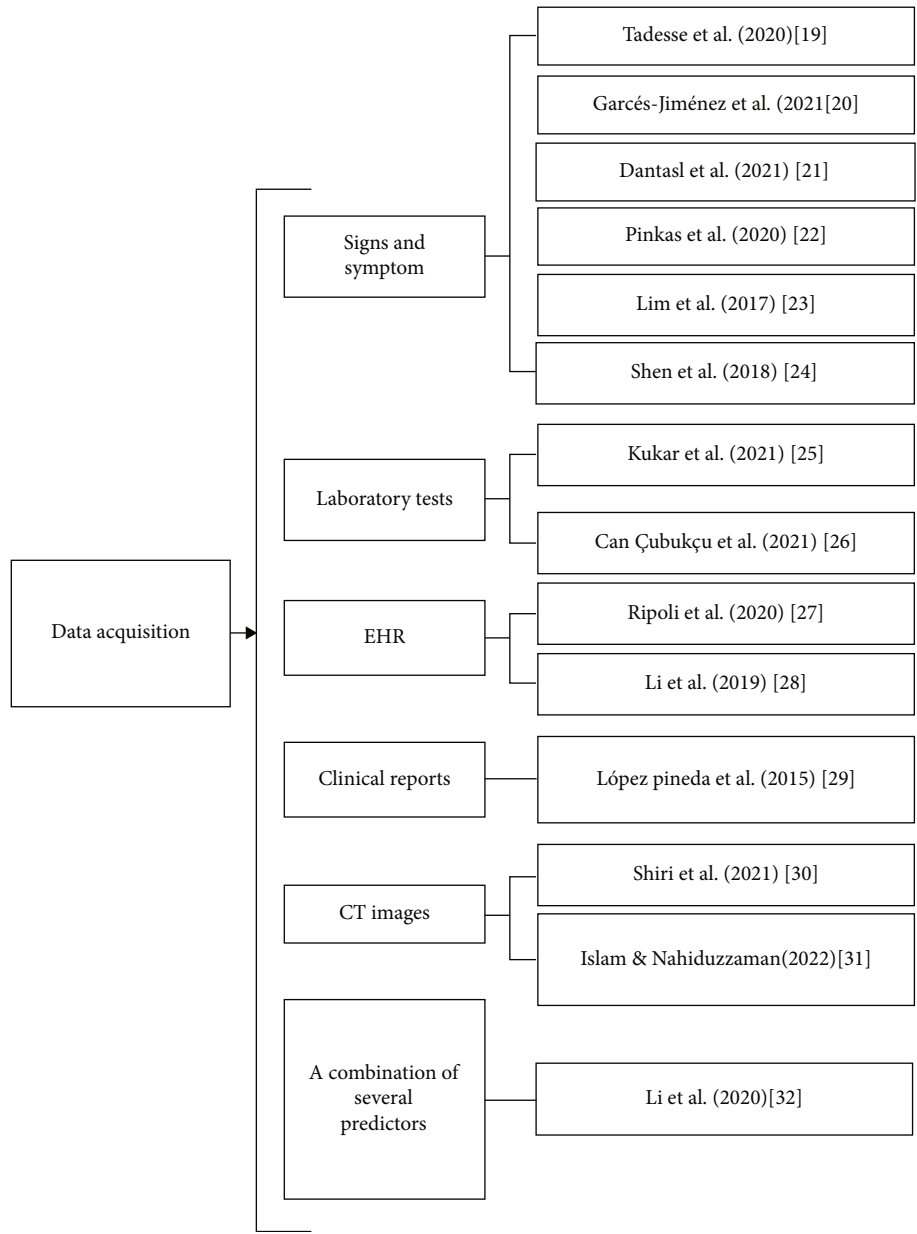

FIGURE 2: Data acquisition methods for the 14 selected articles.

A support vector machine (SVM) was utilized as a classifier by [19, 22, 26] to diagnose various infectious diseases. The study [19] experimented with linear and Gaussian kernels for the SVM to classify autonomic nervous system dysfunction (ANSD) levels for HFMD and infectious tetanus diseases. ANSD is the main cause of death for both HFMD and tetanus. The study sought to detect the ANSD level automatically and showed that Gaussian kernels provided the best results. Moreover, it used different measurements such as accuracy, precision, specificity, recall, and F1-score.

However, manual encoding of features is the main limitation of the study.

Four ML models, i.e., SVM, RF, gradient boosting (XGBoost), and LR, utilized complete blood count (CBC) parameter results and clinical chemistry (CC) results to diagnose the COVID-19 infectious disease [26]. However, the RF model performed the best on the present study's dataset, whereas the adopted SVM model performed the best on the external validation dataset.

Moreover, models trained on CBC and CC provided better results than those trained on CBC only. The study reported that ML models could detect COVID-19 when the population has more severe cases, eventually improving sensitivity. In addition, it demonstrated that eosinophil count is the most important feature that the model uses.

TABLE 3: Dataset specifications for the reviewed articles.

| Reference | Size | Data acquisition method | Infectious disease |
|---|---|---|---|
| [19] | Time-series of physiological data | Wearable sensors to collect multivariate physiological data | Tetanus and HFMD |
| [20] | 60 individuals | Medical sensors, hub, and Android-based app to collect vital signs | Skin and soft tissue infection, urinary tract infection, and acute respiratory infection |
| [21] | 49,721 users | App-based symptom tracker | SARS-CoV-2 (COVID-19) |
| [22] | 88 individuals | Cellular phone voice recordings | SARS-CoV-2 (COVID-19) |
| [23] | 37,599 tweets | Social media messages | Latent infectious diseases |
| [24] | 1,317,018 classes, 7,731,914 axioms, and 1,269,340 inheritance relations | Multiple medical ontologies | 507 infectious diseases |
| [25] | 31268 reports | NLP tool (Topaz) to extract influenza-related findings | Influenza |
| [26] | 52,306 patients | Routine blood tests | SARS-CoV-2 (COVID-19) |
| [27] | 1391 patients | Routine laboratory results | SARS-CoV-2 (COVID-19) |
| [28] | 295 patients | Clinical records | Candidemia |
| [29] | 1118 patients | EHR | CDI |
| [30] | 152 patients | Clinical records and chest CT images | SARS-CoV-2 (COVID-19) |
| [31] | 2482 images | CT scan images | SARS-CoV-2 (COVID-19) |
| [32] | 56,081 patients | Historical and real-time data | 25 infectious diseases |

SARS-CoV-2: severe acute respiratory syndrome coronavirus 2; EHR: electronic health record; HFMD: hand, foot, and mouth disease; CDI: *Clostridium* (*Clostridioides*) difficile infection; CT: computed tomography.

Furthermore, a three-stage architecture supported a semisupervised learning approach [22]. The first stage used a transformer and pretrained it on the unlabeled recordings to transform the frame sequences into "transformer embeddings." In the second stage, recurrent neural network (RNN) classifiers were utilized on speaker vocal inputs, and every classifier produced a score for each COVID-19-positive probability. In the third stage, ensemble stacking generated a meta-model, and scores for each classifier were averaged and assembled into a feature vector per speaker. Finally, a linear SVM trained on the feature vectors weighed the scores and predicted the final result. In addition, cross-validation was used to present the performance metrics for the ensemble stacking meta-model.

Besides the machine-learning algorithm, another routine blood test was used to diagnose COVID-19 from 255 different viral and bacterial infections [25]. It compared various ML techniques, such as SVM, RF, neural network (NN), and extreme gradient boosting (XGBoost). The XGBoost model outperformed others with 5333 patients with negative tests and 160 patients with the positive test used for training. It used a tenfold stratified cross-validation testing procedure to evaluate models' performance.

Ensemble learning employs a mixture of algorithms to solve classification or regression problems that cannot be solved with a single ML model [35]. Moreover, a soft voting ensemble learning model (Gaussian naive Bayes (GNB), SVM, decision tree (DT), LR, and RF) was deployed by [31] on features extracted from a CNN model to diagnose COVID-19. The study used 85% of the images to train the proposed model and 15% of the images to test it. Additionally, a confusion matrix evaluates the robustness of each model by determining the accuracy, F1-score, recall, preci-sion, and area under the curve (AUC) and comparing them with previous work implementing the same dataset.

The study in [21] used a combination of symptoms and ML techniques as a screening model that identifies individuals infected with COVID-19. It compared five ML techniques: LR stepwise, RF, naïve Bayes (NB), decision tree using C5.0 (DT), and XGBoost. Moreover, it utilized different data balancing techniques—upsampling, downsampling, the synthetic minority oversampling technique (SMOTE), and random oversampling examples (ROSE)—and resulted in 25 combinations of ML techniques and sampling strategies. The applied dataset was divided into 80% training and 20% testing sets. Moreover, participants with a probability of more than 50% were classified as infected with COVID-19. Finally, the study applied the Matthews correlation coefficient (MCC), specificity, F1-score, sensitivity, a positive predictive value (PPV), and a negative prepredictive value (NPV) to evaluate the results. The XGBoost, LR, and random forest techniques presented the best median MCCs, followed by naïve Bayes (NB) and decision trees.

Unlike supervised ML algorithms, which train a classifier from manually labeled datasets and then train the classifier for self-classifying unlabeled data, unsupervised ML algorithms train a classifier to discover hidden patterns and structures from unlabeled data without target variables [36]. Unsupervised sentiment analysis was proposed in [23], and it followed a bottom-up approach to discover latent infectious diseases instead of existing diseases. The study applied the Topaz natural language processing (NLP) tool, computational linguistics, and text analysis to Twitter messages. It assumed that body parts, symptoms, and pain locations were mentioned in the text, and symptom weighting vectors for each individual and time period were created.

Table 4: Features used in the diagnosis models.

| Reference | Features |
| --- | --- |
| [19] | ECG, PPG, and IP |
| [20] | EDA, SPO2, body temperature, systolic and diastolic blood pressure, and heart rate |
| [21] | Age, gender, fever, nausea, shortness of breath, diarrhea, coryza, cough, myalgia, loss of smell or anosmia, and living with a confirmed case |
| [22] | Age, gender, cough, short speech utterances, counting, and nonspeech voicing |
| [23] | User IDs, timestamps, geospatial information, and textual information of each message |
| [24] | Patient's data, body temperature, infection site, symptoms, and signs |
| [25] | Nasal swab, lab-confirmed flu, influenza-like illness, suspected flu, viral syndrome, myalgias, rhinorrhea, viral, fever, coughing, chills, sore throat, malaise, arthralgia, pneumonia, wheezing, hoarseness, cervical lymphadenopathy, headache, hemoptysis, fatigue, diarrhea, conjunctivitis, dyspnea, anorexia, nausea, chest pain, cyanosis, pain with eye movement, photophobia, and abdominal cramps |
| [26] | Mean corpuscular hemoglobin concentration, eosinophils count, albumin, prothrombin international normalized ratio, prothrombin activity%, eosinophils%, lymphocyte %, monocyte %, gamma-glutamyltransferase, erythrocyte count, creatinine, alkaline phosphatase, leukocyte count, bilirubin total, aspartate aminotransferase, hematocrit, mean platelet volume, hemoglobin, basophils count, glucose, urea, alanine aminotransferase, age, neutrophils pH count, monocyte count, thrombocytes count, mean corpuscular volume, lymphocyte count, sodium in serum, potassium in serum, neutrophils %, mean corpuscular hemoglobin, bilirubin direct, basophils %, and erythrocyte distribution width |
| [27] | Bilirubin, ALT, AST, LDH, CRP, lymphocytes, creatinine, monocytes, neutrophils, red cell distribution width, platelets, eosinophils, mean corpuscular hemoglobin, leukocytes, mean corpuscular volume, hemoglobin, hematocrit, basophils, mean corpuscular hemoglobin concentration, and RBCs |
| [28] | Age, gender, Charlson's score, previous hospital admission, hospital admission from home, hospital admission from long-term care facility, hospital admission from medical wards, hospital admission from surgical wards, sepsis, fever, previous antifungal therapy, previous antibiotic therapy, in-hospital antibiotic therapy, in-hospital MHIA therapy, steroids during hospitalization, in-hospital immunosuppressants, concomitant infection, previous CDI, PICC, NGT, PN, UC, CVC, recent abdominal surgery, recent nonabdominal surgery, coronary heart disease, heart failure, COPD, diabetes, chronic kidney disease, dialysis, liver disease, pancreatitis, peripheral vascular disease, cerebrovascular disease, dementia, hemiplegia, connective tissue disease, peptic ulcer, leukemia or lymphoma, solid cancer, and metastatic cancer |
| [29] | Age, gender, White race, Charlson-Deyo's score, prior CDI, healthcare-associated CD, immunosuppression, solid organ transplant, metastatic cancer, hypertension, congestive heart failure, diabetes mellitus, chronic kidney disease, depression, concurrent antibiotic use, prior fluoroquinolone use, proton pump inhibitor use, fever, systolic blood pressure, mechanical ventilation, sodium, creatinine, albumin, total bilirubin, white blood cell count, hemoglobin, platelets, ribotypes, positive stool toxin by enzyme immunoassay, polymerase chain reaction cycle threshold, 30-day ICU admission, attributable 30-day ICU admission, 30-day colectomy, attributable 30-day colectomy, 30-day mortality, attributable 30-day mortality, and severe CDI |
| [30] | History, demographics, and clinical data (gender, age, BMI, past medical history of comorbidities, weight, height, history of smoking, level of consciousness, initial vital signs, RR, O2Sat, PR, SBP, DBP, and temperature) Laboratory data (aspartate aminotransferase, monocytes, ALT, ALP, total and direct bilirubin, hemoglobin, WBC, pH, PCO2, HCO3, lymphocytes, CRP, Plt, Cr, BUN, PT, PTT,INR, PCT, sodium, potassium, neutrophils, and eosinophils) Radiological data (type of parenchymal abnormality, emphysema, axial and craniocaudal distribution, pleural effusion, and pericardial effusion) Extracted radiomic features (shape features, gray-level run length matrix, neighboring gray tone difference matrix, gray-level size zone matrix, gray-level cooccurrence matrix, gray-level dependence matrix, and first-order statistics) |
| [31] | 100 prominent features |
| [32] | Age, gender, fever, fatigue, cough, WBC count |

ECG: electrocardiogram signal; PPG: photoplethysmogram; IP: impedance pneumography; EDA: electrodermal activity; SPO2: heart beat rate oxygen saturation; LDH: lactate dehydrogenase; CRP: C-reactive protein; RBCs: red blood cells; ALT: alanine aminotransferase; AST: aspartate aminotransferase; CDI: *Clostridium difficile* infection; PICC: peripherally inserted central catheter; NGT: nasogastric tube; PN: parenteral nutrition; UC: urinary catheter; CVC: central venous catheter; COPD: chronic obstructive pulmonary disease; MHIA: microbiome highly impacting antimicrobials; LOS: length of stay; RR: respiratory rate; O2Sat: O2 saturation; PR: pulse rate; SBP: systolic blood pressure; DBP: diastolic blood pressure; BMI: body mass index; WBC: white blood cells; pH: venous blood gas analysis of acidity; PCO2: carbon dioxide concentration; ALP: alkaline phosphatase; HCO3: bicarbonate concentration; Plt: platelet count; Cr: blood creatinine level; BUN: blood urea nitrogen; PT: prothrombin time; PTT: partial thromboplastin time; INR: prothrombin time normalized with the international normalized ratio; PCT: procalcitonin levels; ICU: intensive care unit.

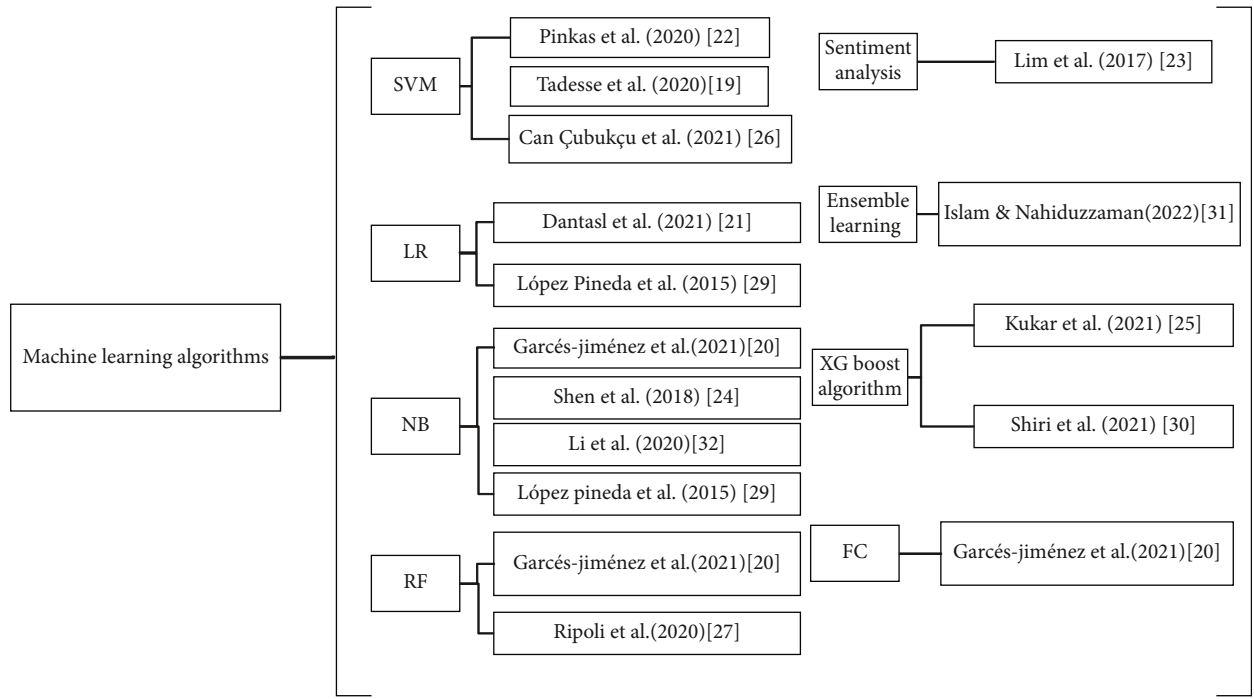

FIGURE 3: ML algorithms used in the reviewed articles.

TABLE 5: Comparison of the reviewed articles concerning the applied ML techniques and the resulted performance.

| Reference | ML technique | Model performance |
|---|---|---|
| [19] | SVM (Gaussian kernels) | For HFMD dataset (AC: 70.9%, precision: 60.6%, SP: 78.0%, F1 score: 55.7%, and recall: 55.9%) |
| | | For tetanus dataset (AC: 80.2%, precision: 78.4%, SP: 53.4%, F1 score: 86.0%, and recall: 98.1%) |
| [20] | NB, filtered classifier (FC), and RF | Success ratio with the weight of each medical variable (vital signs) that affects the prediction |
| [21] | LR | AUC: 0.68, recall: 60%, specificity: 75%, precision: 25%, NPV: 93%, F1 score: 35%, and MCC: 25% |
| [22] | SVM | PFA: 30%, F1 score: 74%, precision: 71%, and recall: 78% |
| [23] | Unsupervised sentiment analysis | Precision: 77.3%, recall: 68%, F1 score: 72.4% |
| [24] | NB | ROC: 89.91%, SN: 47%, and SP: 37% |
| [25] | XGBoost algorithm | AUC: 0.97, SN: 81.9%, SP: 97.9% |
| [26] | SVM | AC: 91.18%, SN: 100%, SP: 84.21%, PPV: 83.33%, F1 score: 90.91%, and AUC: 0.958 |
| [27] | RF | AUC: 0.847, SN: 84.24%, SP: 91%, HLT statistics: 12.779, and HLT $p$: 0.120 |
| [28] | EHR-based model (the study did not mention the used algorithm) | SN: 41.7%, SP: 96.7%, and PPV: 41.7% |
| [29] | NB and LR | AUC: 0.93, BSS: all classifiers achieve positive BSS scores |
| [30] | XGBoost algorithm | AUC: 0.95, AC: 88%, SN: 88%, and SP: 89% |
| [31] | Ensemble learning model | AC: 99.73%, precision: 99.46%, recall: 100%, F1 score: 99.73%, and AUC: 0.9973 |
| [32] | NB | AUC greater than: 98%, SN: 44.44% for hepatitis E and 96.67% for measles, SP: 96.36% for dengue fever and 100% for 5 diseases, median of total accuracy: 97.41%, and $M$-index:0.960 |

SN: sensitivity = recall; AUC: area under the curve; SP: specificity; PPV: positive predictive value; AC: accuracy; PFA: probability of false alarm: BSS: brier skill score; HLT $p$: the $p$ value of the Hosmer–Lemeshow test; ROC: receiver operating characteristic.

The study employed the cooccurrence analysis because there were no training data and to reduce false positives.

On the other hand, seven ML classifiers—Bayesian classifiers (NB, Bayesian network with the K2 algorithm (K2-BN), and efficient Bayesian multivariate classification (EBMC)), function classifiers (LR, SVM, and artificial neural network (ANN)), and decision trees (RF)—were trained using data extracted from ED free-text reports [29]. Their

TABLE 6: Comparison of pros and cons of the reviewed articles.

| Reference | Pros | Cons |
|---|---|---|
| [19] | (i) The proposed method provides efficient hospital resources (ii) Simple and more generic features are used to encode the waveform dynamics in time and frequency domains (iii) Low-cost wearable sensors are used to collect data | (i) The manual encoding of features used to encode the waveform dynamics in time and frequency domains is time-consuming and may have errors (ii) The dataset is small |
| [20] | (i) The study shows an accessible, easy to use, flexible, ubiquitous, and cost-effective eHealth system for diagnosing infectious diseases from vital signs. | (i) The short period of sampling affects the classification results, and more accuracy is needed (ii) The small dataset affects the accuracy of the model |
| [21] | (i) The proposed model shows that a combination of symptoms assisted with the prediction of COVID-19 infection (ii) It helps improve the test strategy by prioritizing users for testing | (i) Some studies criticize the use of symptoms for classifying COVID-19 because of the existence of other respiratory coinfections and the nonspecific nature of some symptoms |
| [22] | (i) The study shows that voice-based screening for COVID-19 is possible (ii) Deep learning is useful in addressing the challenges of the long sequences in voice recordings, uncertain and presumably subtle vocal attributes of early COVID-19, and the lack of large labeled datasets | (i) The study uses a small set of vocal-input types that were self-recorded (ii) The results show a connection between COVID-19 voice symptoms and detection of it, whereas there are no reliable reports to assess this correlation |
| [23] | (i) The study is helpful in diagnosing latent infectious diseases in early stages without prior training data and in a short period. | (i) There is a need to improve the performance of the proposed model and to include accuracy measures when considering social media user information (e.g., age, gender, and posting frequency). |
| [24] | (i) It is more comprehensive compared to other existing works (ii) It establishes a reliable knowledge base for infectious diseases | (i) Symptoms are not weighed to distinguish syndromes and signs (ii) Important etiological factors such as a history of close exposure to other infected patients are not included |
| [25] | (i) The study determines the most useful routine blood parameters for COVID-19 diagnosis from a large number of patients > 5000 (ii) Use of the ML model to diagnose COVID-19 from a routine blood test in the early symptomatic phase is effective when demands on the real-time reverse transcriptase-polymerase chain reaction (RT-PCR) test are enormous | (i) The proposed model might be inefficient at the stage where there are no systemic effects (ii) The study includes only features available from a single center (iii) the positive number of COVID-19 patients is limited in the study. |
| [26] | (i) The proposed ML model can be used as a decision support system tool (ii) A combination of routine laboratory results and ML models improves the COVID-19 diagnosis | (i) The study patients' comorbidities are not available in the dataset (ii) Larger datasets are required (iii) Some important features are not included, such as medical imaging, vital signs, physical examination, symptoms, and increasing sample size (iv) the absence of data pertaining to vaccinations limits the study. |
| [27] | (i) The use of ML to predict Candidemia improves decision-making for appropriateness in antifungal and antibiotic therapies (ii) The use of ML reduces the delay in empirical treatment (iii) The study uses real-world data with a large number of features (as predictors) | (i) There is a need for a large number of patients (ii) It is a retrospective study, and the pooled data are anonymous |
| [28] | (i) The EHR-based model can be used as clinical decision support to predict complicated cases of CDI on the day of diagnosis (ii) Use of the ML approach is feasible for generating accurate and early risk predictions for complicated CDI | (i) The performance metrics that are used are not enough to evaluate the model (ii) The proposed model is not tested in a real-time manner (iii) The obtained results are based on a small dataset from one institution (iv) The study does not mention the applied ML algorithm |

TABLE 6: Continued.

| Reference | Pros | Cons |
|---|---|---|
| [29] | (i) Machine-learning classifiers perform better than expert constructed classifiers using the NLP extraction tool<br>(ii) The large number of ED reports in training classifiers solves the imbalance problem in the dataset<br>(iii) The applied method for dealing with missing values in the study shows improved performance | (i) The study focuses on the data of only one health system<br>(ii) The number of selected features is considered small |
| [30] | (i) The study shows that the combination of clinical data and radiomic features, including all measures in the optimal model, has the height performance, and can effectively predict survival in COVID-19 | (i) The used dataset is small, and there is a lack of an external validation dataset<br>(ii) Clinical studies are required to verify obtained results.<br>(iii) The study tests only one ML classifier and one feature selection method |
| [31] | (i) Diagnosing COVID-19 is faster and more accurate than other traditional methods applied to the same CT image dataset. | (i) None of the COVID-19 variants is included in the study<br>(ii) The proposed model is trained on a tiny dataset<br>(iii) The study uses only a COVID-19 patient dataset to train the model |
| [32] | (i) Various types of predictors are utilized<br>(ii) Data quality is guaranteed in the study<br>(iii) Standard statistical methods are used to validate the model's performance | (i) The validation dataset is available for only 12 out of 25 infectious diseases<br>(ii) The proposed model is limited to only 25 infectious diseases and cannot be generalized<br>(iii) Real-time updates are missing |

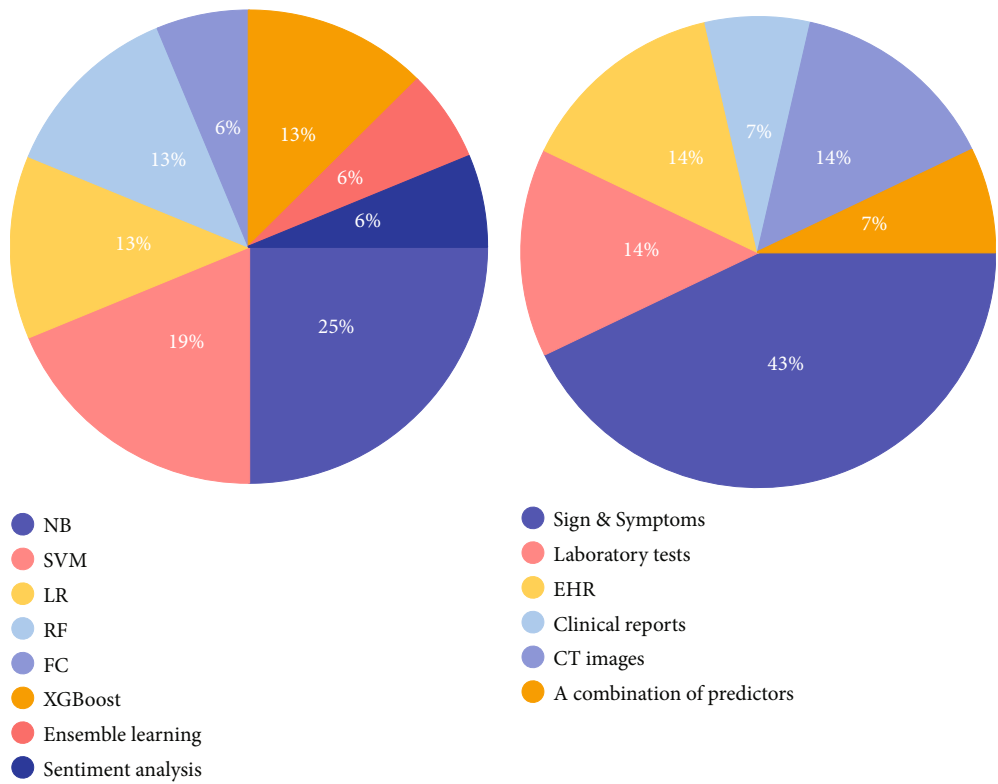

FIGURE 4: Percentages of dataset types and ML algorithms from the reviewed articles.

diagnostic capabilities were compared to an expert-built influenza Bayesian classifier. The NLP results showed a 90% accurate determination of the required clinical findings from ED free-text reports. Missing values in the training dataset were assigned to value (false), whereas in the testing dataset, they were assigned to value missing (M). In the study, only ACU was used to evaluate the performance of the classifiers. In addition, four ML classifiers—NB, SVM, LR, and ANN—obtained the highest AUC, 0.93.

The authors in [20] introduced a combination of ML technology, cloud services, and mobile communications and performed pattern recognition analysis. Three ML algorithms were applied: NB, the filtered classifier (FC), and RF, to allow changes in the weights of the attributes. The 6277 samples from 60 individuals were divided into 95% for training and 5% for validation. It is noticed that heart rate is the most important variable during the classification process. The study implemented NB and FC on a web service and experimented with the weights of the variables through RF as a software application.

In addition, the NB classifier was implemented in [24, 32] to classify several infectious diseases. Moreover, a patient's self-examination is required in [24], and a proposed decision support system that supports infectious disease diagnosis and therapy is constructed. In [24], we focused on the diagnosis part only, which helps to diagnose the type of bacterial infection; it was tested on 84 medical records. The Bayesian classifier possesses high predictive performance in [32], and the output of the model is a list of possible diseases arranged according to the calculated probabilities.

Another combination approach involving the EHR and ML techniques was proposed in [28], whose approach used EHR (4271 features), curated data (23 features) from another study [37], and implemented a combination of both features to train the model. The proposed model in the study was trained on 80% of the data and evaluated on 20%. In addition, it applied the L2 regularization and $k$-best feature selection techniques to control data overfitting and big dimensionality problems, respectively. The results showed that the model based only on EHR outperformed curated and combination-based models in predicting complicated CDI. Moreover, the study implemented the area under the receiver operating characteristic curve (AUROC) to measure the model's discriminative performance.

Furthermore, the RF model was compared to the LR model for early detection of Candidemia in internal medical wards (IMWs) [27]. The RF model was built using both the original features and a shuffled copy of their values. The study used 150 different combinations of the three tuning parameters for the random forest algorithm. The discriminative performance of each model was assessed by sensitivity and specificity values and by the area under the ROC curve (AUC, $C$-statistics). Moreover, the Hosmer–Lemeshow test (HLT) was used to evaluate the models' calibration. Notably, a smaller value of HLT statistics and a greater value of (HLT $p$) for the RF model indicated better performance. The cross-validation procedure showed that the best-tuned RF model outperformed the LR model regarding discrimination and calibration.

A supervised ML algorithm XGBoost classifier was implemented in [30] to train models to find patterns to predict the survival of COVID-19 patients. The predictive models can effectively predict alive or dead status in COVID-19 patients through various combinations of clinical data and radiomic (lung/lesion) features. The dataset was divided into 106 patients for training/validation and 46 for testing to evaluate the selected model.

In addition, bootstrap techniques were used for XGBoost hyperparameter tuning with 1000 repetitions through a random search method. Maximum relevance minimum redundancy (MRMR) was used for feature selection. Table 5 shows the ML techniques used in each article and the performance metrics used to evaluate the developed model.

## 5. Discussion and Future Research Trends

There are numerous limitations reported in diagnostic models of infectious diseases, as illustrated in Table 6. To overcome them, we give some recommendations for future research in the area. Moreover, Figure 4 shows that signs and symptoms were used by several articles to diagnose different infectious diseases, and the most used ML algorithm is NB for different infectious diseases. However, including more critical features can improve models' performance, and the automatic encoding of many features is more efficient than manual encoding. In addition, the consensus in most reviewed articles is the requirement for large dataset to improve accuracy, ensure the reliability of the developed model, and validate results. In addition, diagnostic lab test (s) are required to help confirm disease prognosis.

The ML models could extract delicate prognostic data from routine blood test results that were unobserved by the most experienced clinicians. However, some articles that use a single ML algorithm should focus on evaluating multiple ML algorithms and comparing different models' performance.

Further, NLP is beneficial to extract features from clinical reports, notes, social media, etc. However, a composite of NLP tools may be implemented to increase the accuracy of feature extraction.

Besides, deep learning models, specifically the convolutional neural network (CNN) model, are mighty in extracting huge features from clinical images. These features are then deployed to classify infectious diseases through various ML models. The resulted performance is promised to assist clinicians in diagnosing several infectious diseases from images in their early stages.

EHR contains laboratory data and radiology reports, free-text clinical notes, patients' demographics, etc. Therefore, the use of EHR can achieve improved accuracy. Moreover, if the naïve Bayes classifier is implemented, then calculating the degree of correlation between symptom vectors is essential to evaluate the probability of the infectious disease.

Finally, developing a ML model for detecting and classifying infectious diseases as a web application or API call is a good idea. However, future studies are required to investigate how can ML models perform in real-time. In addition, prospective validations are needed to obtain more robust results. On the other hand, working with a heterogeneous dataset from multiple sources and implementing medical ontologies is more useful to ensure the model's generalizability. Furthermore, the model achieves better discriminative performance as more data becomes available and integrated. Moreover, developing a hybrid model for diagnosing infectious diseases from a vast dataset is recommended. It is important to notice that our review did not perform any meta-analysis because the reviewed data from the studied articles are too heterogeneous [38].

## 6. Conclusion

Infectious diseases can affect humans worldwide and cause death in complicated situations. Many contributions have been developed using various methods to diagnose and classify infectious diseases. Further, ML algorithms can assist in the diagnosis of infectious diseases at early stages. By reviewing the selected articles, we found some limitations in these studies, including small datasets, which is the main limitation. Combining techniques to extract more features is useful and can improve ML predictive models' performance. It is essential to build a smart and generalized health system that can combine medical ontology, real-time heterogeneous data from multiple sources, and the ML predictive model to assist clinicians in diagnosing infectious diseases early. In addition, patients may have access to this system to alert them about possible infectious diseases.

## Conflicts of Interest

The authors declare no conflict of interest.

## Authors' Contributions

All authors have read and agreed to the published version of the manuscript.